\documentclass[letter, 10 pt, conference]{ieeeconf}  
\usepackage{float}
\pdfoutput=1
\usepackage{graphicx}
\usepackage{amssymb}

\usepackage{amsmath}
\usepackage{mathtools}  
\mathtoolsset{showonlyrefs}
\usepackage{algorithmic}
\usepackage{algorithm}
\usepackage{nccmath}
\providecommand{\norm}[1]{\lVert#1\rVert}
\usepackage{amsthm}
\usepackage[usenames,dvipsnames,svgnames,table]{xcolor}
\theoremstyle{plain}
\newtheorem{theorem}{Theorem}

\newtheorem{problem}{Problem}
\newtheorem{assumption}{Assumption}
\usepackage{float}
\floatstyle{plaintop}
\restylefloat{table}
\usepackage{MnSymbol}%
\usepackage{wasysym}%
\providecommand{\norm}[1]{\lVert#1\rVert}
\usepackage{tikz}
\usepackage{verbatim}
\usepackage{graphicx} 
\usepackage{subfigure} 
\usepackage{hyperref}
\hypersetup{colorlinks,allcolors=black}

\usetikzlibrary{positioning}
\tikzstyle{arrow} = [thick,->,>=stealth]

\usetikzlibrary{shapes,shapes.multipart} 
\usetikzlibrary{calc} 

\IEEEoverridecommandlockouts                              
\usepackage{cite} 

\usepackage[utf8]{inputenc}
\usepackage[T1]{fontenc}

\title{\LARGE \bf
Event-Triggered Control for Weight-Unbalanced Directed Robot Networks
}

\author{Juan D. Pabon, Gustavo A. Cardona, Nestor I. Ospina, Juan Calderon, and Eduardo Mojica-Nava.
\thanks{ 
Juan D. Pabon, Nestor I. Ospina, and  Eduardo Mojica-Nava are with  DESYNC-LAB  at Universidad Nacional de Colombia, Bogota, Colombia:\texttt{$\{$jdpabona, niospinag, eamojican$\}$@unal.edu.co}.
Gustavo A. Cardona, is with the AIR-Lab at Lehigh University, PA, USA: \texttt{gcardona@lehigh.edu}. 
Juan M. Calderon is with Universidad Santo Tomas, Bogota, Colombia, and Bethune Cookman University, Daytona, FL.\texttt{juancalderon@usantotomas.edu.co, calderonj@cookman.edu}. }
}

\begin{document}

\maketitle
\thispagestyle{empty}
\pagestyle{empty}

\begin{abstract}
We develop an event-triggered control strategy for a weighted-unbalanced directed homogeneous robot network to reach a dynamic consensus in this work. We present some guarantees for synchronizing a robot network when all robots have access to the reference and when a limited number of robots have access. The proposed event-triggered control can reduce and avoid the periodic updating of the signals. Unlike some current control methods, we prove stability by making use of a logarithmic norm, which extends the possibilities of the control law to be applied to a wide range of directed graphs, in contrast to other works where the event-triggered control can be only implemented over strongly connected and weight-balanced digraphs. We test the performance of our algorithm by carrying out experiments both in simulation and in a real team of robots.
\end{abstract}

\section{INTRODUCTION}
In recent years, the implementation of multi-robotic systems has emerged as a solution for a wide range of applications where more agents can improve the performance and efficiency of a task or process. In addition, robots can communicate between them to optimize the development of a common task \cite{lal2017}.

Two of the most critical tasks in multi-robot systems are navigation and formation control \cite{Gao2018}.  Robots should implement control strategies in a centralized or distributed fashion to achieve the desired behavior. For example, in \cite{blender2016managing}, a centralized control determines the operation of the entire system to perform the seeding in crops. The systems decrease the amount of data that the agents need to process and improve robustness when use distributed approach \cite{ZHOU2018}. Unlike centralized approaches, an effective distributed control technique should consider the communication limitations.
In multi-robot systems, the control is updated according to some rules depending on the information shared by the neighbors. This approach cannot be applied in the continuous-time since robots have to process a sequential algorithm \cite{Wang2019}. The current standard solution is periodic sampling, and synchronous event detection \cite{cao2015event}. However, periodic sampling still has high energy consumption since most systems spend more energy on communication than on computing. 

In this work, we are interested in the event-triggered control of multi-robot systems with directed communication networks. 
With this, we can coordinate the robots teams to travel across fields in which robots need to follow determined trajectories. At the same time, they maintain the desired formation, i.e., seeding task shown in \cite{blender2016managing}. 
By implementing directed graphs, we can reduce communication errors and simplify the communication devices used by robots. In addition, we deal mainly with weight-unbalanced digraphs, graphs that contain nodes in which the sum of the weights of the incoming edges is different from the sum of the outgoing edges.
These graphs represent the communication networks of robot teams in which one-directional communication devices are used and the robots assign different levels of priority to information from various neighbors. 

Some ways to demonstrate the stability of this kind of control approach are presented in \cite{nowzari2014zeno} and \cite{tallapragada2019event}. In contrast, we apply the event-triggered control to a dynamic consensus over a weight-unbalanced directed network. Several works propose the dynamic consensus as it is presented in \cite{Chung2021}, \cite{Kia2015}, and \cite{Kia2019}. In \cite{Chung2021}, the authors deal with the dynamic consensus with event-triggered communication; however, only undirected graphs are considered. In contrast to this work, \cite{Kia2015} deals with the issue of the dynamic consensus considering directed and weight-balanced networks. In this work, the solution considers continuous-time communication. In \cite{Kia2019}, it is presented a solution to the event-triggered control for dynamic consensus over directed graphs. But, it can be only applied in strongly connected and weight-balanced graphs. 

Other control approaches for directed networks are proposed in \cite{Hu2020} and \cite{NOWZARI2016}, where the proposed solutions are Lyapunov-based control strategies, which implies that the communication networks must be strongly connected and weight-balanced. However, the aforementioned works only develop algorithms for weight-balanced digraphs. Furthermore, it imposes the limitation that the robots should assign the same priority level to incoming and outgoing information. In real applications, robots must assign different levels of priority to the information they receive and send. In contrast to these works, the proposed control strategy is based on a differential equations approach. Thus, the conditions of a weight-balanced and strongly connected network are removed. With this approach, we can find the relation between the convergence radius of the network, the applied reference signal, and the event-trigger function.

The main contribution of this paper is threefold. First, we consider a team of differential-drive mobile robots where the system has an external time-varying reference signal injected into the system only through the leader robot. Second, we propose a coordination algorithm that only requires one robot to access a dynamic reference. For this purpose, we analyze the stability of the system using the Logarithmic norm operator \cite{Banasiak2019}, \cite{Perov2017}. Finally, we determine the relation among the time-varying reference signal, the event-trigger condition, and the network connectivity. Unlike previous works, we consider that agents have no access to the reference signal continuously, and only one robot knows this reference. We implement the proposed strategy using the ARGroHBotS \cite{ospina2021argrohbots}.

\textit{Notation}: The notation used for matrices and vectors is $\mathbf{X}$ and $\mathbf{x}$, respectively. The transpose of a matrix $\mathbf{X}$ is $\mathbf{X}^{\top}$ and of a vector $\mathbf{x}$ is $\mathbf{x}^{\top}$. For $a \in \mathbb{R}$, $|a|$ is its absolute value and for a set of numbers $\mathcal{B}$, $|\mathcal{B}|$ is its cardinality. A function $f(\cdot)$ is called a $C^{1}$ function if its derivative exists and is continuous. The operator $\mu(\cdot)$ denotes the logarithmic norm of a matrix, which can take positive and negative real values \cite{Banasiak2019}. 

\section{PRELIMINARIES AND PROBLEM STATEMENT}
\paragraph{Graph Theory}
Consider a weight directed graph $\mathcal{D}=\{\mathcal{V}, \mathcal{E}, \mathcal{W}\}$ consisting of a set of vertices $\mathcal{V}=\{1,\ldots,n\}$, a set of edges $\mathcal{E}$, and a set of weights $\mathcal{W}$. For the directed Graph $\mathcal{D}$, if the ordered pair $(v_{i},v_{j}) \in \mathcal{E}$ exists the edge  $(v_{j},v_{i})$ may or may not exist. The matrix $\mathbf{A}(\mathcal{D})$ represents the adjacency relation between the nodes $v_{i}$ and $v_{j}$, this matrix is defined by $[\mathbf{A}(\mathcal{D})]_{ij}=w_{ij} \in \mathcal{W}$ if the edge $(v_{j},v_{i}) \in \mathcal{E}$ exists  and $[\mathbf{A}(\mathcal{D})]_{ij}=0$ otherwise. If there is a path between any two vertices of the graph $\mathcal{D}$, then $\mathcal{D}$ is called connected. The  in-degree weighted Laplacian matrix of the graph $\mathcal{D}$ is $\mathbf{L}(\mathcal{D}) = \boldsymbol{\Delta}(\mathcal{D}) - \mathbf{A}(\mathcal{D})$, where $\mathbf{\Delta}(\mathcal{D})$ is the diagonal degree matrix of the diagraph. This matrix is defined as $[\mathbf{\Delta}(\mathcal{D})]_{ii}=d_{in}(v_{i})$, where $d_{in}(v)$ is the weighted in-degree of vertex $v$, that is, $d_{in}(v_{i})=\sum_{\{j|(v_{j},v_{i})\in \mathcal{E}(\mathcal{D})\}} w_{ij}$. For connected graphs, $\mathcal{L}$ has only one zero eigenvalue ($\lambda_{1}(\mathcal{D})=0$), and its eigenvalues can be listed in increasing order as $\lambda_{1}(\mathcal{D})<\lambda_{2}(\mathcal{D})\leq\cdot\cdot\cdot\leq\lambda_{n}(\mathcal{D})$.

\paragraph{System Model}
Consider a system of $n$ differential-drive robots with unicycle dynamics. We define the global reference frame as $W=\{X_{w},\ Y_{w},\ \theta_{w}\}$, and to each robot, the local reference frame is defined as $\{X_{r},\ Y_{r},\ \theta_{r}\}$.  If we only consider the motion in the XY-plane, the dynamics of each robot can be expressed as follows,

{\small
\begin{equation}
\left[
\begin{array}{lcr}
v_{i}\\
\omega_{i}\\
\end{array}
\right]
=
\dfrac{1}{a}
\left[
\begin{array}{lcr}
\ a\cos\theta_{i} & \ a\sin\theta_{i} \\
-\sin\theta_{i} & \ \ \cos\theta_{i}\\
\end{array}
\right]
\dot{x}_{i},
\label{roboteq3}
\end{equation}
}
and thus, we now have an equation that let us find the linear and angular velocities of each robot from the desired velocities in $X_{w}$ and $Y_{w}$ axes. This is an equation that lets us map motion from the axes of the robot's local frame to the global reference frame. Therefore, to control the robot team, the dynamics at the team level of the robots can be described by a single integrator. Each robot can compute the agreement state $\mathbf{x}_{i} \in \mathbb{R}^{p}$ in the global reference frame using a coordination algorithm and then using \eqref{roboteq3} to control its dynamics. This fact it is also presented in \cite{bullo2009distributed} and \cite{cortesAndMagunus2017}. Thus, the dynamic of the robots at the team level can be described as follows,  

\begin{equation}
    \mathbf{\dot{x}}_{i}=\mathbf{u}_{i}, \hspace{0.5cm} i \in \mathcal{V}=\{1,\ldots,n\},
    \label{dynamic}
\end{equation}
where $\mathbf{u}_{i} \in \mathbb{R}^{p}$ denotes the control input of the $i$-th robot.

Each robot has a subset $\mathcal{N}_{i} \subset \mathcal{V}$ that represents the rest of the robots which  the $i$-th robot can communicate with. The weight directed graph $\mathcal{D}=(\mathcal{V},\mathcal{E},\mathcal{W})$ consists of a set of terrestrial robots $\mathcal{V}=\{1,\ldots,n\}$, a set of edges $\mathcal{E}=\{(i,j) \in \mathcal{V} \times \mathcal{V} | j \in \mathcal{N}_{i} \}$, and $\mathcal{W}$ that denotes set of weights of the edges meaning the priority of the information coming for each robot, considering that the higher priority will be given to those robots receiving information directly from the leader robot.

We consider the existence of a robot $\mathbf{x}_L$ as the robot which has access to the reference trajectory we want all the robots in $\mathcal{D}$ to follow. With this, only one team robot must have the hardware required to the navigation across the fields following determined trajectories. Other robots only need to reach a consensus to achieve the desired trajectories. The way that the reference trajectory is sent to other robots depends on an event-triggered approach. When a trigger condition executed by the leader is fulfilled, this robot sends its state to the robots belonging to the $\mathcal{N}_L$ set. 
As the full connection of the graph is not granted, the communication between robots occurs in asynchronous time. To generalize the type of possible graphs to implement, we consider weight-unbalanced digraphs.
\begin{problem}
Given a desired trajectory in robot $\mathbf{x}_L$ and a set of robots with dynamic \eqref{dynamic}, which communicate between them according to graph $\mathcal{D}$. The goal is to find control inputs $\mathbf{u}_i$ such that if all of the robots have access to the leader robot, they will follow the leader's trajectory and determine when it is necessary to update the control depending on a trigger condition.
\end{problem}

It is worth noting that not always all the robots can have access to the leader robot $\mathbf{x}_L$, as a consequence of greater distances that the communication system cannot handle or just by design in the algorithm, among others. We formulate a second problem tackling this approach.
\begin{problem}
 Given a desired trajectory in robot $\mathbf{x}_L$ and a set of robots with dynamic \eqref{dynamic}, which communicate between them according to graph $\mathcal{D}$, the goal is to find control inputs $\mathbf{u}_i$ when not all of the robots have access to the leader robot, showing that they are able to move describing the leader's trajectory, and determine when is necessary to update the control depending on a trigger-condition.
\end{problem}
To study both cases, we design a control law and then some stability properties are shown. 
\section{Control Design}
\label{controlDesign}
This section shows the design of the control strategy to coordinate a multi-robot system described by a weight-unbalanced directed network. The control law considers two cases. In the first case, all robots have access to a reference signal. The second case considers that only some robots have access to this signal. The main purpose of the control law is to achieve a dynamic consensus and synchronize the network with the reference while each robot applies a distributed event-triggered control. To prove the stability of the systems under the action of the event-triggered control, an upper bound is found for the norm of the disagreement matrix of the system. Thus, we can guarantee that all robots of the network converge to a ball around the desired behavior.

\paragraph{Design of Control Law}

Consider $n$ robots with states $\mathbf{x}_{i} \in \mathbb{R}^{p}$, where $i=1,\ldots, n$, $\mathbf{x}_j \in \mathbb{R}^p$, and $j \in \mathcal{N}_i$, being $\mathcal{N}_i$ the set containing the in-neighbors of the $i$-th robot, and $\mathbf{x}_{L} \in \mathbb{R}^{p}$ the reference signal applied to the network. We propose the following distributed event-triggered control algorithm 

\begin{equation}
    \mathbf{\dot{x}}_{i}(t)=-\sum_{j\in \mathcal{N}_i}(\mathbf{\hat{x}}_{i}- \mathbf{\hat{x}}_{j}) - \alpha m_{i}(\mathbf{\hat{x}}_{i}- \mathbf{\hat{x}}_{L}),
    \label{controlLaw}
\end{equation}
where $\mathbf{\hat{x}}_{i}, \mathbf{\hat{x}}_{j}$, and $\mathbf{\hat{x}}_{L}$ are the last sample states of robot $i$, robot $j$, and reference robot, respectively. Additionally, $\alpha \in \mathbb{R}_{>0}$ is a design parameter, and $m_{i}\in \{0,1\}$ indicates if the $i$-th robot has access to the reference, being equals to $m_i=1$ if access is granted or $m_{i}=0$ otherwise. A robot's state change depends on its in-neighbors' relative state and the difference between the robot state and the reference if the robot has accessed it. We define the difference between the last sampled state of robot $i$ and the current state as the sampling error as follows 
\begin{equation}
    \mathbf{e}_{i}(t)=(\mathbf{\hat{x}}_{i} - \mathbf{x}_{i}(t))^{\top}.
    \label{samplingError}
\end{equation}

 An event-trigger condition $g(t)$ executed on each robot determines when robots need to transmit their current states to their out-neighbors. Hence, a robot performs \eqref{controlLaw} with the information of the last sampled state of its in-neighbors and the reference. In this case, the external robot that applies the reference signal also implements an event-trigger condition generating that the reference signal to be applied in the network when the reference robot determines it. As a consequence, the intervention of the reference signal in the network is asynchronous.

The disagreement matrix of the network is defined as
\begin{equation}
    \boldsymbol{\delta}=\mathbf{X}-\mathbf{1}_{n}\otimes \mathbf{x}_{L},
    \label{disagreementMatrix}
\end{equation}
where $\mathbf{X}=(\mathbf{x}_{1}^{\top}, \ldots, \mathbf{x}_{n}^{\top})^{\top}$ is a matrix that contains the information of all of the states of the robots, $\mathbf{1}_n \in \mathbb{R}^n$ is column vector of ones, and $\otimes$ is the Kronecker product. The matrix $\boldsymbol{\delta} \in \mathbb{R}^{n \times p}$ contains in each column the difference between the robot's states and the reference signal.  The derivative of this matrix is equivalent to
\begin{equation}
\boldsymbol{\dot{\delta}}=\mathbf{\dot{X}}-\mathbf{1}_{n}\otimes \mathbf{\dot{x}}_{L}.
\label{derivativeDisagreementMatrix}
\end{equation}

Using \eqref{samplingError} and \eqref{disagreementMatrix}, the control law \eqref{controlLaw} can be expressed in stack vector form as follows 
{\footnotesize	
\begin{equation}
    \mathbf{\dot{X}}  =-(\mathbf{L}\otimes \mathbf{I}_{p})\mathbf{X} -(\mathbf{L}\otimes \mathbf{I}_{p})\mathbf{e} - (\alpha \mathbf{M}\otimes \mathbf{I}_{p})\boldsymbol{\delta} - (\alpha \mathbf{M}\otimes \mathbf{I}_{p})(\mathbf{e}-\mathbf{1}_{n}\otimes \mathbf{e}_{L}),
    \label{vectorForm}
\end{equation}
}
where the term $\textbf{e} \in \mathbb{R}^{n\times p}$ is the matrix that contains the sampling error of all robots. The vector $\mathbf{e}_{L} \in \mathbb{R}^{1\times p}$ is the sampling error of the reference robot, $\mathbf{L}\in \mathbb{R}^{n \times n}$ is the Laplacian matrix of the graph, $\mathbf{I}_p \in \mathbb{R}^{p \times p}$ is an identity matrix, and $\mathbf{M} \in \mathbb{R}^{n\times n}$ is the matrix that indicates which robots have access to the reference robot. Solving $\mathbf{\dot{X}}$ from \eqref{derivativeDisagreementMatrix} and replacing it in \eqref{vectorForm} we obtain
{\small
\begin{equation}
\begin{split}
    \boldsymbol{\dot{\delta}}  = & -(\mathbf{L}\otimes \mathbf{I}_{p})\mathbf{X} -(\mathbf{L}\otimes \mathbf{I}_{p})\mathbf{e} - (\alpha \mathbf{M}\otimes \mathbf{I}_{p})\boldsymbol{\delta} - (\alpha \mathbf{M}\otimes \mathbf{I}_{p})\mathbf{e}\\
    & + (\alpha \mathbf{M}\otimes \mathbf{I}_{p})(\mathbf{1}_{n}\otimes \mathbf{e}_{L}) - \mathbf{1}_{n}\otimes \mathbf{\dot{x}}_{L}.
\end{split}
\label{previous}
\end{equation}
}

From the definition of the disagreement matrix, $\mathbf{X}=\boldsymbol{\delta} +\mathbf{1}_{n}\otimes \mathbf{x}_{L}$. Thus, \eqref{previous} can be expressed as follows
{\footnotesize
\begin{equation*}
\begin{split}
    \boldsymbol{\dot{\delta}} = & -(\mathbf{L}\otimes \mathbf{I}_{p})\boldsymbol{\delta} -(\mathbf{L}\otimes \mathbf{I}_{p})(\mathbf{1}_{n}\otimes \mathbf{x}_{L}) -(\mathbf{L}\otimes \mathbf{I}_{p})\mathbf{e} - (\alpha \mathbf{M}\otimes \mathbf{I}_{p})\boldsymbol{\delta} \\
    & - (\alpha \mathbf{M}\otimes \mathbf{I}_{p})\mathbf{e} + (\alpha \mathbf{M}\otimes \mathbf{I}_{p})(\mathbf{1}_{n}\otimes \mathbf{e}_{L}) - \mathbf{1}_{n}\otimes \mathbf{\dot{x}}_{L}.
\end{split}
\end{equation*}
}
Note that from definition of the Laplacian matrix {\small$-(\mathbf{L}\otimes \mathbf{I}_{p})(\mathbf{1}_{n}\otimes \mathbf{x}_{L})=0$}. With the foregoing we obtain
{\small
\begin{equation}
\begin{split}
    \boldsymbol{\dot{\delta}} + ((\mathbf{L}+\alpha \mathbf{M})\otimes \mathbf{I}_{p})\boldsymbol{\delta} = & -((\mathbf{L}+\alpha \mathbf{M})\otimes \mathbf{I}_{p})\mathbf{e} \\
    & + (\alpha \mathbf{M}\otimes \mathbf{I}_{p})(\mathbf{1}_{n}\otimes \mathbf{e}_{L}) - \mathbf{1}_{n}\otimes \mathbf{\dot{x}}_{L}.
\end{split}
\label{differentialEquation}
\end{equation}
}
The disagreement matrix behavior with the sampling errors and the reference signal change is stated in \eqref{differentialEquation} which can be solved to test the stability of the network.

\paragraph{Stability Analysis}
\label{stabilityAnalysis}
From the solution of \eqref{differentialEquation}, we can study the behavior of the network when the control law \eqref{controlLaw} is applied. This is the main result of our work: to guarantee the stability of a weight-unbalanced network while robots implement the distributed event-triggered control. This result is stated in the following theorem.
\begin{assumption}
The event-trigger condition $g(t) \in \mathbb{R}>0$ is a continuous and bounded function.
\end{assumption}

\begin{theorem}
Consider a weight-unbalanced directed network with the control law \eqref{controlLaw} applied. If the event-trigger condition has the form $f(\norm{\mathbf{e}_{i}(t)})=g(t)$ and the norm of the reference signal derivative is lower than the upper-bound $c_{r}$, then the asymptotic stability of the network is guaranteed when the Logarithmic norm operator $\mu (\cdot)$ of the matrix $-((\mathbf{L}+\alpha \mathbf{M})\otimes \mathbf{I}_{p})$ is negative. 
\end{theorem}

\textit{Proof :} To observe the stability of the network, we can solve \eqref{differentialEquation} using the integrating factor method, which is given by
\begin{equation*}
    {e}^{\int(\mathbf{L}+\alpha \mathbf{M})\otimes \mathbb{I}_{p} dt} = {e}^{(\mathbf{L}+\alpha \mathbf{M})\otimes \mathbf{I}_{p}t}.
\end{equation*}
Let's $\mathbf{A}=-(\mathbf{L}+\alpha \mathbf{M})\otimes \mathbf{I}_{p}$, solving \eqref{differentialEquation}, then
{\scriptsize
\begin{equation*}
    \boldsymbol{\delta}= \boldsymbol{\delta}(0)e^{\mathbf{A}t} + \int_{0}^{t}e^{\mathbf{A}(t-s)}(\mathbf{A}\mathbf{e}(s) + (\alpha \mathbf{M}\otimes \mathbf{I}_{p})(\mathbf{1}_{n}\otimes \mathbf{e}_{L}(s)) - \mathbf{1}_{n}\otimes \mathbf{\dot{x}}_{L}(s))ds.
\end{equation*}
}
Taking the Euclidean norm on both sides of the previous equation, applying triangle inequality, and Kronecker product properties, we can express this equation as follows


\begin{equation*}
\begin{split}
    \norm{\boldsymbol{\delta}} \leq & \norm{\boldsymbol{\delta}(0)}\norm{e^{\mathbf{A}t}} + \norm{\mathbf{A}}\int_{0}^{t}\norm{e^{\mathbf{A}(t-s)}}\norm{\mathbf{e}(s)}ds\\  
    & +\alpha \norm{\mathbf{M1}_{n}}\int_{0}^{t}\norm{e^{\mathbf{A}(t-s)}}\norm{\mathbf{e}_{L}(s)}ds\\
    & +\norm{\mathbf{1}_{n}}\int_{0}^{t}\norm{e^{\mathbf{A}(t-s)}}\norm{\mathbf{\dot{x}}_{L}(s)}ds.
\end{split}
\end{equation*}

Let $f(\norm{\mathbf{e}_{i}(t)})=g(t)$ be the event-trigger condition with $g(t) \in \mathbb{R}>0$. If $g(t)$ is a bounded function, then for all $t\geq 0$ we have that $\norm{\mathbf{e}}\leq\sqrt{n}g(t)$. If the reference robot sends its state to other robots when its sampling error satisfies the same trigger condition, then $\norm{\mathbf{e}_{L}}\leq g(t)$. In addition, if the reference signal is a $C^{1}$ function and its first derivatives for each component are bounded, the next inequality holds for all $t$, $\norm{\mathbf{\dot{x}}_{L}}\leq c_{r}$. With the sampling errors and the norm of the derivative of the reference signal bounded, now we can denote the norm of the disagreement matrix as 
{\footnotesize
\begin{equation*}
\begin{split}
    \norm{\boldsymbol{\delta}} \leq & \norm{\boldsymbol{\delta}(0)}\norm{e^{\mathbf{A}t}} \\
    & + ((\sqrt{n}\norm{\mathbf{A}} + \alpha \norm{\mathbf{M1}_{n}} )g(t)
    +\sqrt{n}c_{r})\int_{0}^{t}\norm{e^{\mathbf{A}(t-s)}}ds.
\end{split}
\end{equation*}
}
Depending on matrix $\mathbf{A}=-((\mathbf{L}+\alpha \mathbf{M})\otimes \mathbf{I}_{2})$, the term $\norm{e^{\mathbf{A}t}}$ may or may not be bounded. Therefore, it is necessary finding a bound for this term to determine if the system is stable or not. Since $t\geq 0$ and the term $(t-s)\geq 0$, using the Logarithmic norm operator properties of the matrix $\mu(\mathbf{A}) \in \mathbb{R}$, we have $\norm{e^{\mathbf{A}t}} \leq e^{\mu(\mathbf{A})t}$ and $\norm{e^{\mathbf{A}(t-s)}} \leq e^{\mu(\mathbf{A})(t-s)}$.

Thus, we obtain
{\scriptsize
\begin{equation*}
      \norm{\boldsymbol{\delta}}  \leq  \norm{\boldsymbol{\delta}(0)}e^{\mu(\mathbf{A})t}
     + ((\sqrt{n}\norm{\mathbf{A}} + \alpha \norm{\mathbf{M1}_{n}} )g(t) 
    +\sqrt{n}c_{r})\left(\dfrac{e^{\mu(\mathbf{A})t} - 1}{\mu(\mathbf{A})}\right).
\end{equation*}}
The Logarithmic norm has a special characteristic; this can be positive or negative. If we can guarantee that $\mu(\mathbf{A})$ is negative, we have that 
{\scriptsize
\begin{equation}
\begin{split}
  \norm{\boldsymbol{\delta}} \leq & \norm{\boldsymbol{\delta}(0)}e^{-|\mu(\mathbf{A})|t}\\
  & + ((\sqrt{n}\norm{\mathbf{A}} + \alpha \norm{\mathbf{M1}_{n}} )g(t) 
    +\sqrt{n}c_{r})\left(\dfrac{1 - e^{-|\mu(\mathbf{A})|t}}{|\mu(\mathbf{A})|}\right).
\end{split}
\label{disagreementNorm}
\end{equation}
}
Thus, we have that $\norm{\boldsymbol{\delta}(t)}$ is bounded for all $t\geq 0$. Therefore we have that the network converges to a neighborhood or radius $\epsilon$ around $\mathbf{x}_{L}$, such that
{\scriptsize
\begin{equation}
    \epsilon  =\lim\limits_{t \to \infty}(\norm{\boldsymbol{\delta}(t)}),
     =\dfrac{1}{|\mu(\mathbf{A})|} \left((\sqrt{n}\norm{\mathbf{A}} + \alpha \norm{\mathbf{M1}_{n}})g(t) + \sqrt{n}c_{r} \right ),
    \label{convergenceRadius}
\end{equation}
}
proving in this way that the system is stable and can synchronize with the reference signal. \hspace{2.7 cm} $\blacksquare$

\textbf{Remark 1. } The time constant {\scriptsize $\tau=\dfrac{1}{|\mu(\mathbf{A})|}$}, of the exponential term in \eqref{disagreementNorm} determines the settling time of the bound of {\scriptsize$\norm{\boldsymbol{\delta}(t)}$}. Therefore, the convergence time of the network is less or equal than the settling time of the upper-bound.

Until now, the stability of the network depends on the sign of $\mu(\mathbf{A})$. Therefore, to guarantee that the aforementioned results are valid, it must be guaranteed that the Logarithmic norm is negative. This condition entirely depends on network topology and connections between nodes and the reference. Although to calculate the logarithmic norm, it is necessary to have global information of the network and the connections between agents. It is important to highlight that this is only necessary for the stability analysis and not for the execution of the control strategy. The control law is executed in a distributed way, and each robot requires only local information.

\paragraph{Estimation of Logarithmic Norm}
By properties of the logarithmic norm operator \cite{Banasiak2019}, \cite{Perov2017}, if a matrix is negative definite, then its logarithmic norm is negative. Therefore, for both cases of study, when all or some robots have access to the reference signal, we need to guarantee that $\mathbf{A}=-((\mathbf{L}+\alpha \mathbf{M})\otimes \mathbf{I}_{2})$ is a negative definite matrix. It is guaranteed if $\mathbf{L} + \alpha \mathbf{M}$ is a positive definite matrix. $\mathbf{L}$ is the Laplacian matrix of the graph composed by the follower robots.

According to \cite[Theorem 2.12]{Magnus2010}, if we choose the connections among the reference robot and other robots such that in the directed graph $\mathcal{D}$ formed by all robots, including the reference robot, this robot is chosen as the root of $\mathcal{D}$. The connections among robots are such that there are no directed cycles in the digraph. Then, the principal minor of the Laplacian matrix of $\mathcal{D}$ ($\mathbf{L}_{1}(\mathcal{D})$) is a positive definite matrix. Thanks to the structure of the proposed control law, we have that $\mathbf{L} + \alpha \mathbf{M} = \mathbf{L}_{1}(\mathcal{D})$. Thus, the matrix $\mathbf{L} + \alpha \mathbf{M}$ is positive definite. Therefore $\mathbf{A}=-((\mathbf{L}+\alpha \mathbf{M})\otimes \mathbf{I}_{2})$ is a negative definite matrix. With this, we can guarantee that $\mu(\mathbf{A})$ is negative.




\paragraph{Exclusion of Zeno behavior}
To guarantee that the event-triggered control law \eqref{controlLaw} with event-trigger condition $g(t)$ is properly designed, the trigger-condition must generate a limited number of events in a determined time. It is the exclusion of the Zeno behavior. In this case, the event-trigger condition implemented by each robot depends on its state. It can be found a minimum inter-event time or MIET (minimum time between two consecutive events $t_{k}$ and $t_{k+1}$) greater than zero, which guarantees the exclusion of the Zeno behavior. The MIET for any robot $i$ can be expressed as $\text{MIET} = t^{i}_{k+1} - t^{i}_{k}$. 
Since the event-trigger condition applied by each robot depends on the distance since the last time that the robot sends information to its out-neighbors (sampling error), we can express the MIET of the robot $i$ in terms of its velocity and the distance that determines the threshold of the event-trigger condition as follows $\text{MIET} =g(t)/v_{i\max}$. Since $g(t)>0$ and a physical system can not reach an infinity speed ($v_{i\max}$ is bounded), the MIET is always greater than zero. Therefore the exclusion of the Zeno behavior is guaranteed.

\section{Simulation and Implementation Results}
In this section, we illustrate the theoretical results of the previous sections through numerical simulations and implementation on real robotic platforms.

\pgfdeclarelayer{background}
\pgfsetlayers{background,main}

\tikzstyle{vertex}=[circle,fill=black!25,minimum size=20pt,inner sep=0pt]
\tikzstyle{selected vertex} = [vertex, fill=red!24]
\tikzstyle{edge} = [draw,thick,-]
\tikzstyle{weight} = [font=\small]
\tikzstyle{selected edge} = [draw,line width=5pt,-,red!50]
\tikzstyle{ignored edge} = [draw,line width=5pt,-,black!20]

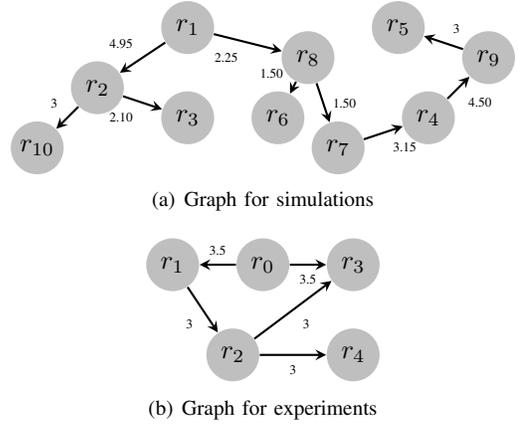
\begin{figure}[]
\begin{center}
\begin{subfigure}[Graph for simulations]{

\begin{tikzpicture}[scale=0.8, auto,swap]    
    \foreach \pos/\name in {{(-1,0)/r_{10}}, {(0,1)/r_{2}}, {(1.5,0.5)/r_{3}},
                            {(1.5,2)/r_{1}}, {(3,0.5)/r_{6}}, {(3.5,1.5)/r_{8}}, {(4,0)/r_{7}}, {(5,2)/r_{5}},
                            {(5.5,0.5)/r_{4}}, {(6.5,1.5)/r_{9}}}
        \node[vertex] (\name) at \pos {$\name$};
        \draw [arrow] (r_{1}) edge node[] {{\tiny  4.95}}  (r_{2});
        \draw [arrow] (r_{2}) edge node[] {{\tiny 2.10}}  (r_{3});
        \draw [arrow] (r_{2}) edge node[] {{\tiny 3}}     (r_{10});
        \draw [arrow] (r_{1}) edge node[] {{\tiny 2.25}}  (r_{8});
        \draw [arrow] (r_{8}) edge node[] {{\tiny 1.50}}  (r_{6});
        \draw [arrow] (r_{8}) edge node[right] {{\tiny 1.50}}  (r_{7});
        \draw [arrow] (r_{7}) edge node[] {{\tiny 3.15}}  (r_{4});
        \draw [arrow] (r_{4}) edge node[] {{\tiny 4.50}}  (r_{9});
         \draw [arrow] (r_{9}) edge node[] {{\tiny 3}}  (r_{5});
         
\end{tikzpicture}
}

\end{subfigure}

\begin{subfigure}[Graph for experiments]{

\begin{tikzpicture}[scale=0.8, auto,swap]    
    \foreach \pos/\name in {{(0,2)/r_{1}}, {(1,0.5)/r_{2}}, {(1.5,2)/r_{0}},
                            {(3,2)/r_{3}}, {(3,0.5)/r_{4}}}
        \node[vertex] (\name) at \pos {$\name$};
        \draw [arrow] (r_{0}) edge node[] {{\tiny  3.5}}  (r_{1});
        \draw [arrow] (r_{0}) edge node[] {{\tiny  3.5}}  (r_{3});
        \draw [arrow] (r_{1}) edge node[] {{\tiny   3}}   (r_{2});
        \draw [arrow] (r_{2}) edge node[] {{\tiny   3}}   (r_{3});
        \draw [arrow] (r_{2}) edge node[] {{\tiny   3}}    (r_{4});

\end{tikzpicture}
}

\end{subfigure}
\end{center}
\caption{Graphs implemented by the robots teams in simulation and experimental tests.}
\label{grafo}
\end{figure}

\paragraph{Simulation Results}
Here two cases are exposed. First, all robots in Fig.\ref{grafo} have access to the leader robot. Second, only robots $1, 2, 5,$ and $9$ have access to the leader. In both cases, only the leader robot knows the reference and the followers' robots reach the dynamic consensus to compute their movements. The following reference signal is applied

\begin{figure}[]
\begin{center}
    \includegraphics[width=0.4\textwidth]{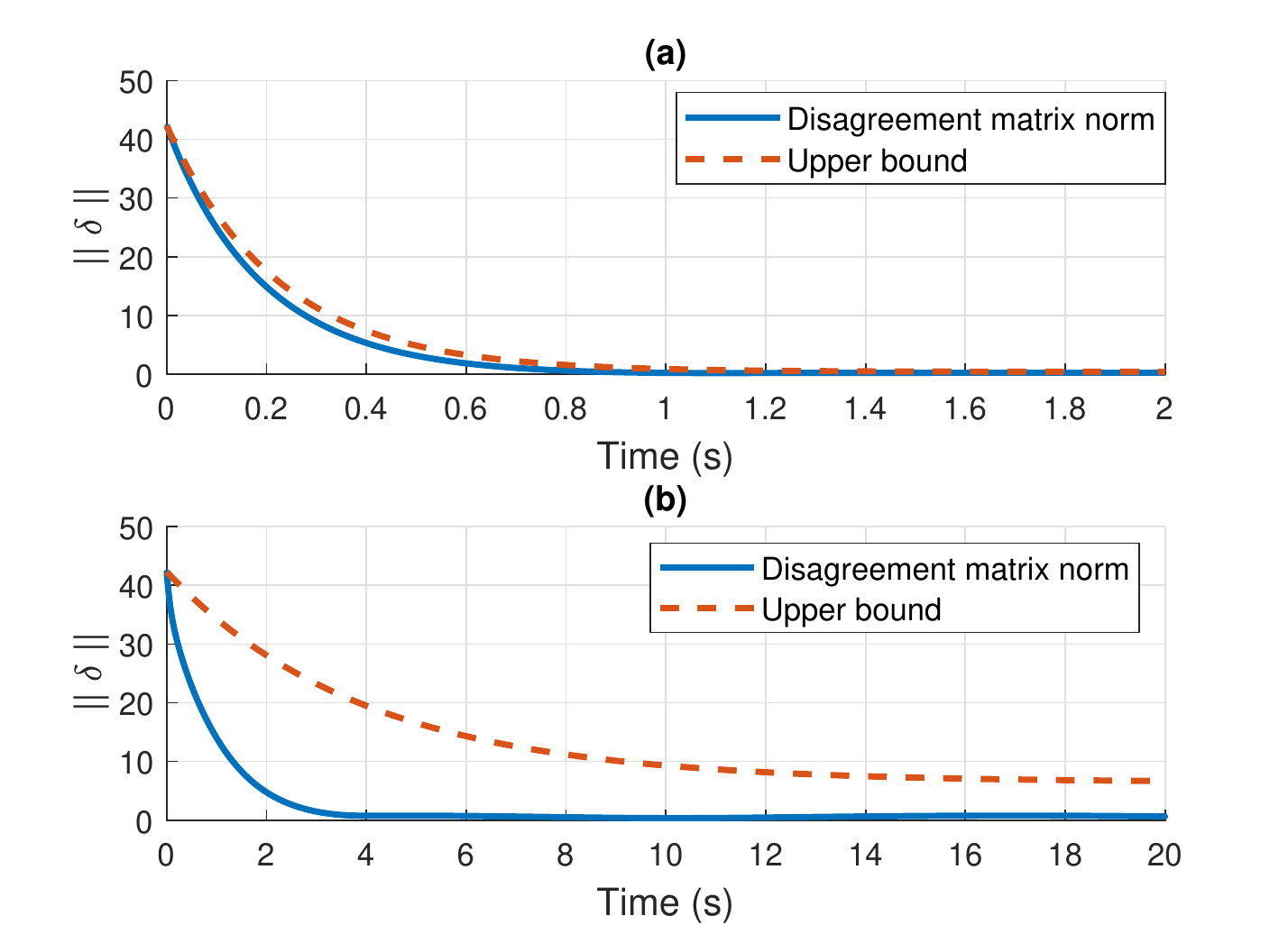}
    \caption{Disagreement matrix norm. The norm represents the radius of the ball determining the convergence region of the system. a) all robots have accesses to reference and b) only some robots have accesses to the reference.}
    \label{norm}
\end{center}
\end{figure}

{\tiny
\begin{equation*}
\mathbf{x}_{L}
=
\left[
\begin{array}{lcr}
\left(\sin(0.2t)e^{0.007t} + 5e^{-0.001t}+0.7\sin(0.04t)+0.07t\right)e^{0.005t}\\
\left(\sin(0.2t)e^{0.007t}+10e^{-0.001t}+0.7\sin(0.04t)\right)e^{-0.005t}
\end{array}
\right],
\end{equation*}
}
where $\norm{\mathbf{\dot{x}}_{L}}\leq 0.43$. The designed control law \eqref{controlLaw} was applied in the first case with $\alpha=2.5$ and in second case $\alpha=5$, resulting in $\mu(\mathbf{A}_{1})=-1.97$ and $\mu(\mathbf{A}_{2})=-0.25$, respectively. For both cases, all robots employ the following event-triggered condition
$\norm{\mathbf{e}_{i}(t)}=0.01 + 0.01e^{\mu(\mathbf{A})t}$.
When the norm of the sampling error reaches the upper bound, robots transmit their states. This event-trigger condition is the same applied by the reference robot. Thus, this robot injects the reference signal into the network in its event times. Using \eqref{convergenceRadius}, it is obtained that the convergence radius in each case is $\epsilon_{1}=1.1$ and $\epsilon_{2}=9.2$. In Fig.\ref{norm} are shown the results. It can be seen that the norm of the disagreement matrix is bounded by \eqref{disagreementNorm} in both cases. When only four robots have access to the reference, the upper bound is greater than when all robots have access. 
Increasing the number of robots when all robots are connected to the leader remains almost invariant to the system behavior. But, when the number of robots disconnected from the leader increase, the poor connectivity with the leader has a strong effect on the system behavior. 
It causes that the logarithmic norm decreases as the ratio between disconnected and connected robots to the leader increases since the eigenvalues of matrix $\mathbf{A}$ decrease. It produces a smaller magnitude of the logarithmic norm, which causes that the convergence radius $\epsilon$ in \eqref{convergenceRadius} and the settling time $\tau$ of the systems increase.

Due to the event-trigger condition and the characteristics of the reference signal, the time between two consecutive event-times (times in which a robot transmits its state) changes when time passes. This is caused by the asynchronous communication generated by the event-triggered control. It can be seen in the next section when the inter-event times are analyzed. 

As shown in the simulation results, we can demonstrate that the use of the logarithmic norm allows us to find the conditions under which the systems for the two cases of the study were stable. Moreover, the use of this operator allowed easily finding an upper bound to the settling time of the multi-robot system.

Using \eqref{convergenceRadius}, we can observe the relation among network topology, the trigger-condition, and the speed of the leader robot, which determines the convergence radius of the robots around the leader. When few agents have access to the leader robot, it is observed that the upper bound in the logarithmic norm increase. Despite this, the system can reach a convergence radius close to the leader. With these results, we have all the necessary tools to determine the behavior of a multi-robot system in which the proposed distributed event-triggered algorithm is implemented. 

\paragraph{Implementation on Real Robots}
To test the control algorithm on physical robots, we use a team of five differential-drive robots. One of these robots has the reference signal, and each other use communication to reach the reference. For the applied control law \eqref{controlLaw} in the experiment we set $\alpha=3.5$. The event-trigger function implemented by each robot is $\norm{\mathbf{e}_{i}(t)}=0.1 + 25e^{\mu(\mathbf{A})t}$. With this, each robot computes the desired velocities in the global reference frame and then implements \eqref{roboteq3} to find the linear and angular velocities in its local reference frame.

With the foregoing, each time that a robot travels a distance equal to $0.1 + 25e^{\mu(\mathbf{A})t}$ cm from the last time that an event-time occurred for this robot, the trigger-function triggers the next event-time. According to Fig. \ref{grafo}b, where  $r_{0}$ represents the reference robot, the value of $\mu(\mathbf{A})$ with $A=-((\mathbf{L}+\alpha \mathbf{M})\otimes \mathbf{I}_{2})$ in this case is equal to $\mu(A)=-0.77$. The experiment is shown below. 

In the experiment, a sinusoidal trajectory was implemented. the trajectory established in the reference robot was $x=20\cos(0.018t)+10$ in the horizontal axis and $y=0.14t$ in the vertical axis for $0 \leq t \leq 49$\ s. The maximum speed reached by the reference robot in this trajectory was equal to $c_{r}=20$ cm/s. To study the performance of the robots, we used the upper bound for the norm of the disagreement matrix, which allows us to see if the system holds the condition imposed by \eqref{disagreementNorm} and \eqref{convergenceRadius}. 

Moreover, since the control law makes the robots reach the same position, to avoid collisions and determine the desired geometric formation of the robots required for the tasks as the seeding, the local frameworks of the robots are separated 60 cm between them on the horizontal axis. With this, the robots will describe the shape of the trajectory described by the leader but displaced 60 cm among them. Therefore, we use a local framework where the reference point starts at the initial conditions. 


\begin{figure}[]
\begin{center}
    \includegraphics[width=0.45\textwidth]{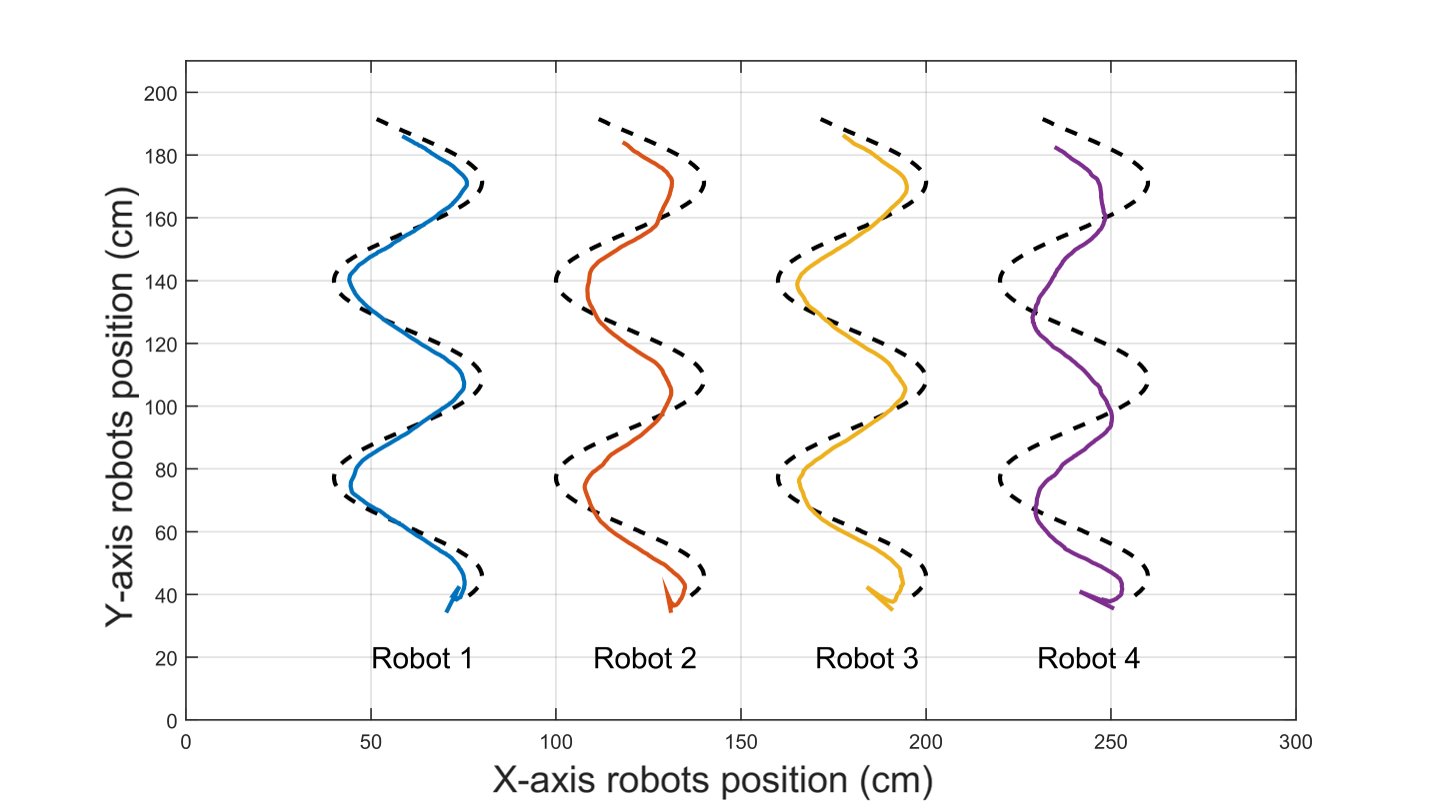}
    \caption{Trajectories described by the robots in the sinusoidal reference. \url{https://youtu.be/mvQeEzIGbgQ}}
    \label{zigzagPos}
    \vspace{-0.15 cm}
\end{center}
\end{figure}
\begin{figure}[]
\begin{center}
    \includegraphics[width=0.42\textwidth]{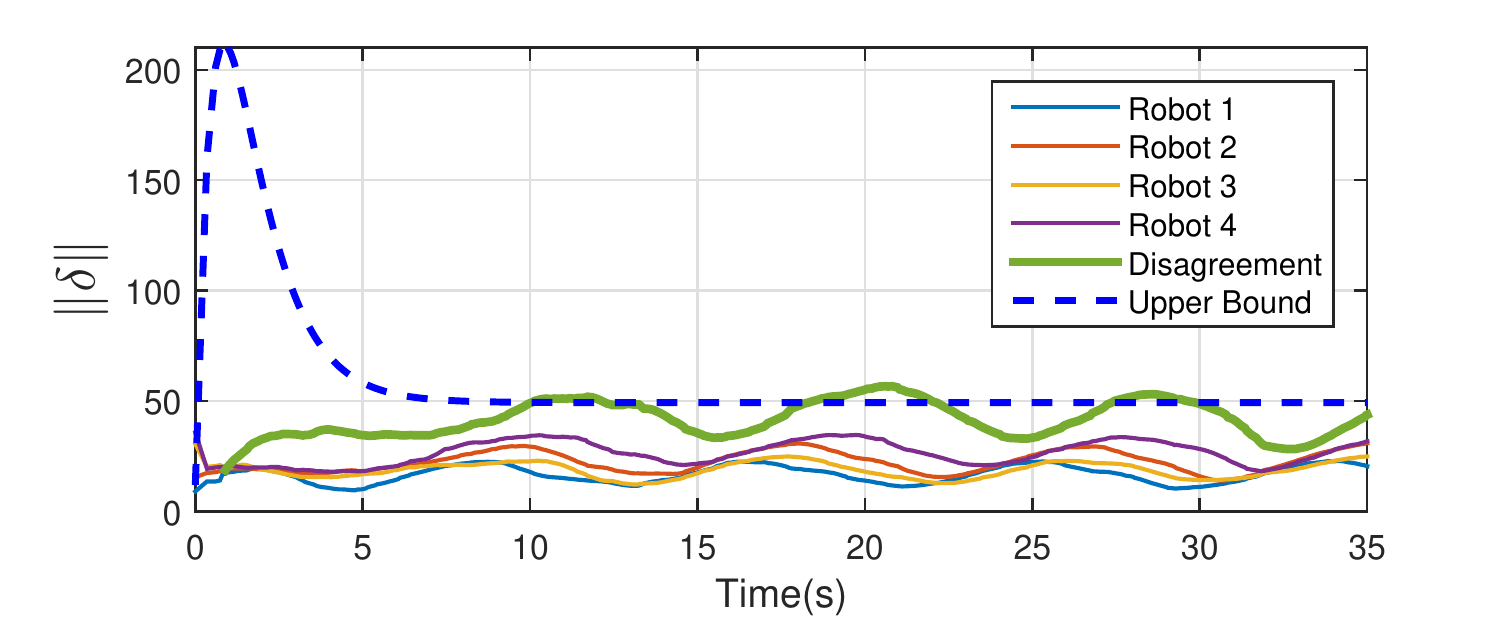}
    \caption{Norm of the disagreement matrix.}
    \label{zigzagDis}
    \vspace{-0.15 cm}
\end{center}
\end{figure}

It can be seen in Fig. \ref{zigzagPos} that robots are close to the desired trajectory described by dashed lines. Robots 2 and 4 do not have direct access to the reference robot; therefore, the trajectories that they follow are deformed. It could be solved by connecting these robots to the leader or increasing the connections weight with their in-neighbors. According to the stability analysis performed in Section III.b,  
the norm of the disagreement matrix of the system must be bounded by \eqref{disagreementNorm}. It is the main performance factor to analyze the experimental results. We can see in Fig. \ref{zigzagDis} that most of the time, the norm of the disagreement matrix is lower than the bound imposed by \eqref{disagreementNorm}. The times in which the disagreement matrix norm exceeds the bound are caused by the dynamics of the robots. 

The time in which robots can execute the control algorithm is 100 ms; if the robot reaches the trigger condition, it could move for this amount of time before taking some control action. Also, after taking control actions, robots need some time to reach the position determined by the coordination algorithm. Thus, when the sampling error of some robot reaches the trigger condition does not mean that the robot can execute a control input immediately. It causes that sometimes the disagreement matrix norm exceeds the upper bound. However, the response of the multi-robot system is governed by this bound.      
The event-times are generated any time that the trigger-function of a robot determines if it is necessary to send information to the out-neighbor robots. These events are no periodic, and they are determined by the movement of the robots and by the trigger-function that they implement. Since the event-trigger function of each robot is a decreasing exponential function whose initial value is much larger than the final value, in the first seconds, the inter-event times are greater than in the rest of the time. It allows the robots to reduce the number of times that they need to communicate with other robots. Also, in the parts of the trajectory where the velocity decreases, robots can decrease the times that they need to send information to their out-neighbors.      

\section{CONCLUSION}
We develop a control strategy based on a stability analysis with differential equations. It guarantees that a multi-robot system converges to the value of the reference signal and allows estimating how long it takes. This can be applied to weight-unbalanced directed robot networks. Furthermore, it provides excellent flexibility since it removes the restriction that most event-triggered controllers for directed networks impose and the necessity to operate in a weight-balanced communication network. In experimental results, it is observed that the convergence radius of the network is directly related to the degree of connectivity among the leader robot and the follower robots. 


\bibliographystyle{ieeetr}
\bibliography{referencias}
\end{document}